\documentclass{article}
\usepackage{spconf,amsmath,graphicx}

\usepackage{hhline}
\usepackage{multirow,amssymb}
\usepackage{hyperref}

\title{Semantic Role Aware Correlation Transformer \\ for Text to Video Retrieval}

\name{Burak Satar$^{\star \ast}$ \qquad Zhu Hongyuan$^{\star}$ \qquad Xavier Bresson$^{\dagger}$ \thanks{This research is supported by the Agency for Science, Technology and Research (A*STAR) under its AME Programmatic Funding Scheme (Project A18A2b0046). Xavier Bresson is supported by the NRF Fellowship NRFF2017-10. Correspondence to \href{zhuh@i2r.a-star.edu.sg}{Dr.Zhu Hongyuan}.}
\qquad Joo Hwee Lim$^{\star \ast}$
}
\address{\{burak\_satar, zhuh, joohwee\}@i2r.a-star.edu.sg, xavier@nus.edu.sg\\$^{\star}$ Institute for Infocomm Research, A*STAR, Singapore \\ $^{\ast}$ School of Computer Science and Engineering, NTU, Singapore \\ $^{\dagger}$ Department of Computer Science, National University of Singapore } 

\begin{document}

\maketitle

\begin{abstract}
With the emergence of social media, voluminous video clips are uploaded every day, and retrieving the most relevant visual content with a language query becomes critical. Most approaches aim to learn a joint embedding space for plain textual and visual contents without adequately exploiting their intra-modality structures and inter-modality correlations. This paper proposes a novel transformer that explicitly disentangles the text and video into semantic roles of objects, spatial contexts and temporal contexts with an attention scheme to learn the intra- and inter-role correlations among the three roles to discover discriminative features for matching at different levels. The preliminary results on popular YouCook2 indicate that our approach surpasses a current state-of-the-art method, with a high margin in all metrics. It also overpasses two SOTA methods in terms of two metrics.  
\end{abstract}


\section{Introduction}
\label{sec:intro}
With the popularity of TikTok, Youtube and Instagram, millions of videos are uploaded every minute, making video retrieval an important function for users to find relevant content. Conventional models are based on keywords query \cite{Chang2015SemanticCD, Habibian2014CompositeCD}. However, unstructured keywords are limited and insufficient to retrieve fine-grained and compositional events. Therefore, communities are shifting attention to cross-modal video text retrieval \cite{Chen_2020_CVPR, mithun2020, miech18learning,miech19howto100m, miech20endtoend} which can retrieve videos using natural language descriptions that includes more structured details.

Most existing cross-modal text video retrieval maps each modality into a joint embedding space to measure their similarities. Some works \cite{Liu2019a, mithun2020, dong_cvpr19} directly embeds whole videos and texts into flat vectors for matching, however losing fine-grained details in texts and videos. To avoid losing those details, some other works \cite{song2019polysemous, Yu_2018_ECCV} align a sequence of frames and words in texts to compute overall
similarities. Although these approaches have achieved certain progress, aligning video and text is still an open problem, given a huge semantic gap between video and text. 

Recently, there are different attempts to resolve the video-text semantic gap. \cite{Chen_2020_CVPR} propose decomposing text into three semantic roles (events, actions and entities) and then embedding 2D video features into these three spaces accordingly for matching. Another line of research uses a BERT-like transformer \cite{NIPS2017_3f5ee243,gabeur2020mmt, devlin-etal-2019-bert} to learn the text-video correspondence, based on recent mixture-of-expert embedding \cite{miech18learning} which requires large-scale dataset for pre-training. 

\begin{figure*}[htb]
\centering
\centerline{\includegraphics[width=\textwidth]{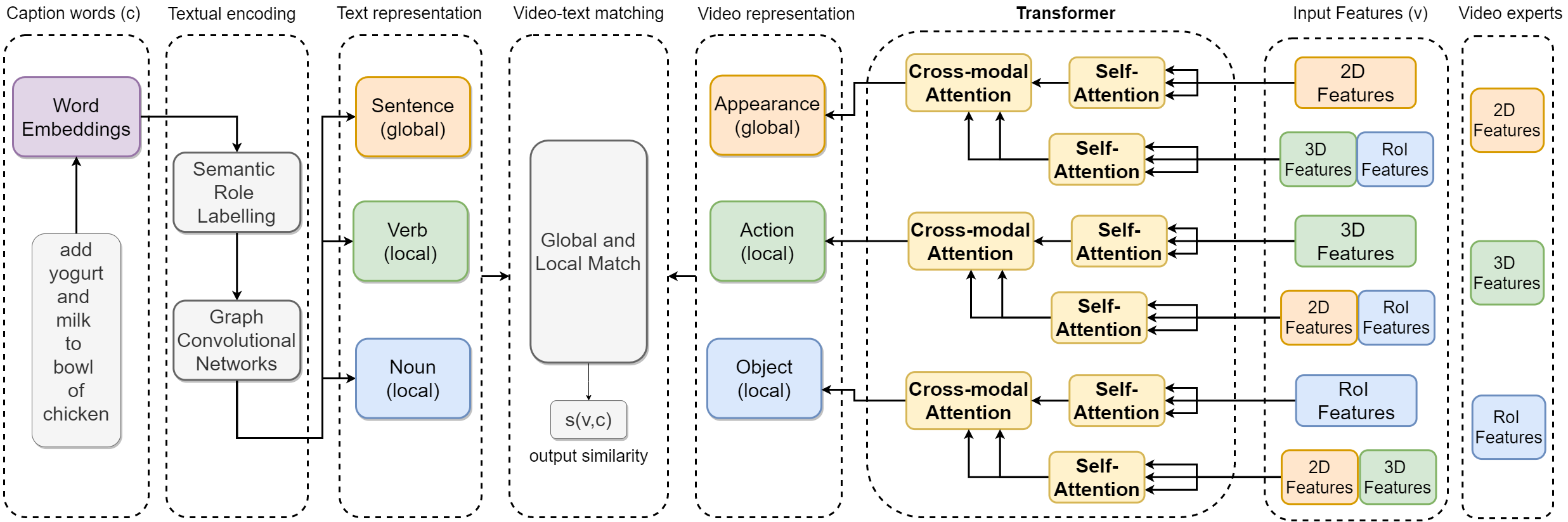}}

\caption{Overview of our model on text-to-video retrieval, where the textual input is a caption of a video clip, and the visual input are the visual expert features from RoI, 2D Frames and 3D clips. It computes the similarity between the caption and a candidate video clip in the three embedding spaces correspond to object contexts, spatial contexts and temporal contexts by embracing transformer with self-attention and cross-modal attention that can capture the specific and complement information within three semantic role modalities. In the cross-modal matching unit, a matching score is calculated for three-level of embeddings. Then, we average the similarities and utilize contrastive ranking loss as a training objective. }
\label{fig:res2}
\end{figure*}

We propose a novel transformer architecture for video-text matching inspired by \cite{Chen_2020_CVPR} and \cite{NIPS2017_3f5ee243,gabeur2020mmt, devlin-etal-2019-bert}. Different from \cite{Chen_2020_CVPR}, which only considers multi-head embedding of the spatial frame and ignores the interaction between different visual contexts, our method explicitly considers more fine-grained visual encoding of object contexts, spatial contexts and temporal contexts by embedding RoI regions, 2D frames and video sequences into the corresponding space with their interactions. Different from \cite{NIPS2017_3f5ee243,gabeur2020mmt, devlin-etal-2019-bert}, which only uses self-attention to discover modality-specific information, our method uses a self-attention scheme to discover modality-specific discriminative features.  Moreover, our model utilizes cross-modal attention to consider the interactions between object, spatial and temporal contexts to discover modality-complement features for better align video and text.

We experimented with the YouCook2 dataset for the text-to-video retrieval task. The results show that our approach surpasses a recent SOTA with a high margin. While our approach gives better results over other approaches in terms of two parameters(R@1, R@5), it falls a close behind them in terms of the other two parameters(R@10, MedR). 

\section{Related Work}
\label{sec:relatedwork}

The text-video retrieval task is challenging because there are significant semantic and structure gap between videos and text. Mithun \textit{et al.} \cite{mithun2020} employ multimodal image, motion, audio modalities in video. Liu \textit{et al.} \cite{Liu2019a} further utilize all modalities that can be extracted from videos such as speech contents and scene texts for video encoding. Dong \textit{et al.} \cite{dong_cvpr19} use biGRU and CNN to encode sequential videos and texts. Yu \textit{et al.} \cite{Yu_2018_ECCV} propose to fuse the sequential interaction of videos and texts for video text retrieval. Song \textit{et al.} \cite{song2019polysemous} employ to align encoded video and text elements for matching. Tan \textit{et al.} \cite{tan2020wman} use LSTM and Graph Neural Networks to capture cross-modal relation. Chen \textit{et al.} \cite{Chen_2020_CVPR} disentangle phrases into different part-of-speech such as verbs and nouns for fine-grained retrieval, and also it uses GNNs to encode the interactions between the roles in text modalities. Our work complements Chen \textit{et al.} \cite{Chen_2020_CVPR} by focusing on video encoding parts using a transformer to learn interactions between object contexts, spatial contexts and video contexts. 

Some methods \cite{videobert, Zhu_2020_Actbert} extend BERT-like models with self-supervised learning to have better video-text representation. For example, while Sun \textit{et al.} \cite{videobert} build upon the BERT model to learn bidirectional joint distributions over sequences of visual and linguistic tokens. Zhu \textit{et al.} \cite{Zhu_2020_Actbert} aim to encode global actions, local regions, objects, and linguistic descriptions with self-supervised learning without considering the linguistic and visual structures. Our work aims to exploit transformer like architecture to learn interactions for better object contexts, spatial context and video context encoding. 

Most recent papers \cite{miech18learning, gabeur2020mmt} use complex datasets such as \cite{miech19howto100m} for pre-training to have a more general textual and visual embeddings. Alayrac \textit{et al.} \cite{alayrac2020selfsupervised} even use an audio-based dataset \cite{audioset} to have audio embedding.  In this current work, we have yet to exploit pre-training on a huge dataset motivated by studying the computational time and the effects of pre-training absence, although our work can incorporate pre-trained backbones.

\section{Method}
\label{sec:method}

Figure \ref{fig:res2} is an illustration of our semantic role aware correlation transformer model, which includes three main blocks: textual encoding, visual encoding, and cross-modal matching.

\subsection{Textual Encoding}
\label{sec:methodtext}

Text description internally has a hierarchical structure. For example, the whole sentence can describe global contexts; nouns and verbs can define object and actions. We use an off-the-shelf toolkit \cite{shi2019simple} for semantic role labelling to parse these components and the semantic relationship between them in a graph structure. Verb nodes are directly connected with the global node by showing temporal relations of various actions. Noun nodes are connected with verb nodes by defining the objects. $r_{ij}$ edges show the semantic relationship between verb nodes $i$ and object nodes $j$. For graph representation, we follow the method proposed by Chen \textit{et al.} \cite{Chen_2020_CVPR}. This approach, which is shown in Eq. \ref{eq_text}, uses factorized weights in GCN, where $W_t$ is a transformation matrix and shared among all relationship types. $W_t$ denotes a unique matrix for different semantic roles. $g_a$ and $g_b$ denote node embeddings which can be for sentences, verbs or objects. $\beta$ is the outcome after attention is applied at the nodes. 

\begin{equation}
g^l_{i} = g^{l-1}_i + \sum_{j\varepsilon  N_i}(\beta_{ij} (W^l_t \odot W_rr_{ij})g_j)
\label{eq_text}
\end{equation}

\subsection{Visual Encoding}
\label{sec:methodvisual}

Parsing videos into hierarchical semantic features is more challenging than texts since it includes detection, action segmentation and so on.  We extract spatial context features $F_S$ with 2D CNNs by ResNet-152 \cite{2D_CNNs}, pre-trained on Imagenet \cite{imagenet}. The temporal context features $F_T$ are extracted with 3D CNNs by ResNeXt-101 \cite{3D_CNNs} pre-trained on Kinetics \cite{kinetics}, then we perform temporal max-pooling on feature maps. We extract object context features $F_O$ by using Faster R-CNN \cite{faster_rcnn}, pre-trained on MS COCO \cite{lin2015microsoft}, and the backbone is ResNet-101 \cite{2D_CNNs}. We follow the settings of Zhu \textit{et al.} \cite{Zhu_2020_Actbert} by extracting the features at 1 FPS after RoI-pooling happens. Confidence threshold is set as 0.4, while each frame could contain up to ten boxes. Dimension for all three expert embedding is 2048.

To capture the modality-complement and specific information, we adapt a transformer with attention modules \cite{NIPS2017_3f5ee243,gabeur2020mmt, devlin-etal-2019-bert, MDVC_Iashin_2020}. For example, to generate spatial context-specific feature $s_e$, the temporal context features $F_T$, and object context features $F_O$ are concatenated in feature dimension and fed into an encoder with a multi-head self-attention layer and a feed-forward linear layer FF, as shown in Eq. \ref{encoder}.

\begin{equation}
\begin{aligned}
&f_e = \textrm{Concat}(F_T,F_O)& \\
&z_e = \textrm{Norm}\big(\textrm{MultiHead}(f_e, f_e, f_e) + f_e\big)& \\ 
&s_e = \textrm{Norm}\big(\textrm{FF}(z_e) + z_e\big)& \\
\label{encoder}
\end{aligned}  
\end{equation}

The attention layers are multi-headed as Vaswani \textit{et al.} \cite{NIPS2017_3f5ee243}'s dot-product attention, and each layer also follows layer normalization and residual connection, as shown in Eq. \ref{transformer}. All the W matrices are trainable parameters. 

\begin{equation}
\begin{aligned}
&\textrm{MultiHead}(Q,K,V) = \textrm{Concat}(\textrm{Head}_1, ..., \textrm{Head}_h)W^O \\ 
&\textrm{Head}_i = \textrm{Attention}(QW_i^Q, KW_i^K, VW_i^V) \\ 
&\textrm{Attention}(Q,K,V) = \sigma\Big(\frac{QK^T}{\sqrt{d}}V\Big)
\end{aligned}
\label{transformer}
\end{equation}

Eq. \ref{decoder} shows the formula of generating spatial context-complement feature. The spatial context feature $F_S$ is first fed into the self-attention layer to encode into $z_s$, then with cross-modal attention is conditioned on $s_e$ and $z_s$ to generate complement feature $c_e$. Next, it is encoded through feed-forward linear layer FF and together with the modality-specific feature $s_e$ to deliver final embeddings for the spatial context $E_S$. 'Norm' refers to layer normalization. 

\begin{equation}
\begin{aligned}
&z_s = \textrm{Norm}\big(\textrm{MultiHead}(F_S, F_S, F_S) + F_S\big)& \\
&c_e = \textrm{Norm}\big(\textrm{MultiHead}(s_e, z_s, z_s) + z_s\big)& \\
&E_S = \textrm{Norm}\big(\textrm{FF}(c_e) + c_e\big)& \\
\label{decoder}
\end{aligned}  
\end{equation}

The same process is applied to generate the final embeddings $E_T$ and $E_O$ for temporal context feature $F_T$ and object context feature $F_O$, respectively, for matching. 

\subsection{Cross-modal Matching}
\label{sec:methodmatch}

We utilize cosine similarity to calculate the cross-modal matching score for each level by corresponding visual and textual embeddings.

\begin{equation}
s(V,C) = \frac{<v, c>}{||v||_2 ||c||_2 }
\end{equation}

Then, we average the similarities and utilize contrastive ranking loss as a training objective. Our aim is to have such positive pair $(v_p,c_p)$ pushed away from the negative pairs $(v_p,c_n)$ and $(v_n,c_p)$ than a set margin. $v$ and $c$ refer to the video and textual representation, respectively. This matching applies to three levels of embeddings. $\Delta$ represents the pre-defined margin, which is for training with contrastive loss.

\begin{equation}
\begin{aligned}
L(v_p,c_p) = [\Delta + s(v_p, c_n) - s(v_p,c_p)] + \\ [\Delta + s(v_n, c_p) - s(v_p,c_p)]
\end{aligned}
\end{equation}

\begin{table*}[htb!]
\centering
\caption{Text-to-video retrieval comparison with SOTA approaches on YouCook2 validation set. 'Visual Backbone' only refers to 3D CNNs Features. Our method surpasses the SOTA methods in the first two parameters when without pre-training.}
\label{tab:comparison}
\begin{tabular}{|c|c|c|c|c|c|c|c|}
\hline
\textbf{Method} & \textbf{Pre-training} & \textbf{Visual Backbone} & \textbf{Batch Size} & \textbf{R@1↑} & \textbf{R@5↑}  & \textbf{R@10↑} & \textbf{MedR↓} \\ \hline
Random          & No                    & -                        & -                   & 0.03         & 0.15          & 0.3           & 1675          \\ 
Miech et al \cite{miech19howto100m}     & No                    & ResNeXt-101              & -                   & 4.2          & 13.7          & 21.5          & 65            \\ 
HGLMM \cite{hglmm}           & No                    & -                        & -                   & 4.6          & 14.3          & 21.6          & 75            \\ 
HGR \cite{Chen_2020_CVPR}            & No                    & ResNeXt-101              & 32                  & 4.7          & 14.1          & 20.0          & 87            \\ 
\textbf{Ours}            & No                    & ResNeXt-101              & 32                  & \textbf{5.3} & \textbf{14.5} & 20.8          & 77            \\ \hline \hline
Miech et al+FT \cite{miech19howto100m}  & HowTo100M             & ResNeXt-101              & -                   & 8.2          & 24.5          & 35.3          & 24            \\ 
ActBert \cite{Zhu_2020_Actbert}        & HowTo100M             & ResNet-3D                & -                   & 9.6          & 26.7          & 38.0          & 19            \\ 
MMV FAC \cite{alayrac2020selfsupervised}        & HowTo100M+AudioSet    & TSM-50                   & 4096                & 11.5         & 30.2          & 41.5          & 16            \\ 
MIL-NCE \cite{miech20endtoend}        & HowTo100M             & S3D                      & 8192                & 15.1         & 38.0          & 51.2          & 10            \\ \hline
\end{tabular}
\end{table*}

\section{Experiments}

We compare our model with state-of-the-art methods on text-to-video retrieval task, shown in Table \ref{tab:comparison}. We also share an ablation study using various expert embeddings in different hierarchical levels, shown in Table \ref{tab:ablation}.

\subsection{Dataset and Metrics}

\textbf{Dataset.} We evaluate our model on YouCook2 \cite{ZhXuCoAAAI18}. It is a video dataset on cooking gathered from YouTube. The videos have a diverse number of cooking styles and methods. It includes 89 recipe types and 14k video clips correlated by imperative English captions defining the action. Note that since annotations for the test set is not published yet, we evaluate the task on validation clips which is around 3.5k totally. 

\textbf{Evaluation metrics.} The task is retrieving video clips based on text queries. We evaluate our model using common metrics on recall at various sets and median rank. R@1, R@5, R@10 gives the number of accurately retrieved clips in the ranking list's top associated rates.  MedR gives the median rank of correct clips in the ranking list. While higher is better for recall metrics, lower is better for the median rank.

\begin{table*}[ht]
\centering
\caption{Ablation studies on YouCook2 dataset to investigate the contributions of various feature experts at different levels. The same ablation is also done on HGR method \cite{Chen_2020_CVPR} since it is a strong baseline. On 2D + 3D visual features setting, when the feature dimension is 4096, concatenation is done on dimension one; otherwise is done on dimension zero. Our model surpasses HGR with the same hierarchical features with a high margin by using cross-modal attention.}
\label{tab:ablation}
\begin{tabular}{|c|c|c|c|c|c|c|c|c|}
\hline
\multirow{2}{*}{\textbf{Method}} & \multicolumn{3}{c|}{\textbf{Visual Features}}           & \multirow{2}{*}{\textbf{\begin{tabular}[c]{@{}c@{}}Feature\\ Dimension\end{tabular}}} & \multirow{2}{*}{\textbf{R@1↑}} & \multirow{2}{*}{\textbf{R@5↑}} & \multirow{2}{*}{\textbf{R@10↑}} & \multirow{2}{*}{\textbf{MedR↓}} \\ \cline{2-4}
                                 & \textbf{Appearance} & \textbf{Action} & \textbf{Object} &                                                                                       &                               &                               &                                &                                \\ \hline
HGR \cite{Chen_2020_CVPR} : Ours                       & 2D                  & 2D              & 2D              & 2048                                                                                  & 4.7 : 4.2                     & 13.8 : 13.7                   & 19.7 : 19.4                    & 86 : 86                        \\ 
HGR \cite{Chen_2020_CVPR} : Ours                       & 2D + 3D             & 2D + 3D         & 2D + 3D         & 2048                                                                                  & 4.8 : 4.5                     & 14.0 : 13.2                   & 20.3 : 20.0                    & 85 : 85                        \\ 
HGR \cite{Chen_2020_CVPR} : Ours                       & 2D + 3D             & 2D + 3D         & 2D + 3D         & 4096                                                                                  & 4.8 : 4.5                     & 14.0 : 13.2                   & 20.3 : 20.0                    & 85 : 85                        \\ 
HGR \cite{Chen_2020_CVPR} : Ours                       & 2D                  & 3D              & RoI             & 2048                                                                                  & 4.7 : \textbf{5.3}                     & 14.1 : \textbf{14.5}                   & 20.0 : \textbf{20.8}                    & 87 : \textbf{77}                        \\ \hline
\end{tabular}
\end{table*}

\subsection{Implementation Details}

\textbf{Training.} We use Glove embeddings \cite{pennington-etal-2014-glove} by setting the word embedding size as 300 for the text encoding. We adopt the text encoding approach of \cite{Chen_2020_CVPR} to disentangle the text embeddings. Graph convolutions have two layers, and it outputs features with 1024 dimensions. A linear layer is applied to each visual expert embeddings to transform their dimensions into 1024. We use a cross-modal attention mechanism for each level in visual encoding. For training, the margin is set to $\Delta=0.2$; the epoch is 100 for each experiment. The mini-batch size is 32.   

\subsection{Comparison}

Table \ref{tab:comparison} shows the comparison with SOTA methods. While 'Visual Backbone' in the table only refers to 3D CNNs Features, ResNet-based models are also used in all the models as 2D CNNs backbone. We focus on training only with YouCook2 dataset; however, we also add the methods when using pre-training for reference.  ’FT’ denotes fine-tuning on YouCook2 dataset. We see that the accelerator type, which directly defines the maximum batch size, and pre-training usage affects the result sharply. There are even differences with the same model when training on different accelerators, epoch, and batch sizes; which can be found on corresponding papers. For example, the MIL-NCE method reaches 50\% of its accuracy when trained with less batch size and epochs. Our method surpasses the HGR method \cite{Chen_2020_CVPR} with a high margin for all metrics. We also outperform the other two SOTA methods: Miech et al. \cite{miech19howto100m}, and HGLMM \cite{hglmm}, in terms of the first two parameters. We think that modality-specific and modality-complement features improve accuracy at R@1 and R@5, which are more demanding and useful for real-world applications.

\subsection{Ablation Studies}

Since we adopt our model from HGR \cite{Chen_2020_CVPR}, we implement extensive ablation to show the improvement in our approach on the visual encoding part. We aim to find how the feature expert combination affects the result. Our model falls short when we feed three levels with only 2D features. The same decrease continues with the concatenated 2D and 3D features at dimension zero as well as the concatenated 2D and 3D features at dimension one. However, when we feed them with 2D, 3D, and RoI features respectively, while the HGR model shows a slight decrease, our model reaches a better result with a high margin. This confirms our insight that inter-modal correlation can be exploited with our proposed cross-modal attention mechanism to achieve better results.

\section{Conclusion}
\label{sec:foot}

Retrieving related video on a textual query gets harder since the number of videos on the internet increases. Most works use one joint embedding space for text-to-video retrieval task without fully exploiting cross-modal features. We propose a hierarchical model representing complex textual and visual features with three joint embedding spaces by utilizing self-attention and cross-modal attention to exploit the modality-specific and modality-complement visual embeddings. Our model surpasses a strong baseline with a high margin, and it also overpasses other SOTA methods in R@1, R@5 metrics.

\clearpage
\newpage
\bibliographystyle{IEEEbib}
\bibliography{refs}

\end{document}